\documentclass[conference]{IEEEtran}
\IEEEoverridecommandlockouts

\usepackage[utf8]{inputenc}
\usepackage[T1]{fontenc}
\usepackage[english]{babel}
\usepackage{cite}
\usepackage{amsmath,amssymb,amsfonts}
\usepackage{algorithm}
\usepackage{algorithmic}
\usepackage{graphicx}
\usepackage{textcomp}
\usepackage{xcolor}
\usepackage{booktabs}
\usepackage{array}
\usepackage{multirow}
\usepackage{url}
\usepackage{balance}
\usepackage{tikz}
\usetikzlibrary{arrows.meta,positioning,shapes.geometric,fit,backgrounds,calc}

\def\BibTeX{{\rm B\kern-.05em{\sc i\kern-.025em b}\kern-.08em
T\kern-.1667em\lower.7ex\hbox{E}\kern-.125emX}}

\begin{document}

\title{Hybridization of a Grouping Metaheuristic and Reinforcement Learning for the One-Dimensional Bin Packing Problem}

\author{
\IEEEauthorblockN{
Mostefai Mounir Sofiane,
Tati Youcef,
Badaoui Ikram,
Bousdjira Nadine,
Hasnaoui Sarah,
Zitouni Rania
}
\IEEEauthorblockA{
\textit{Square Two Team -- Optimization Module}\\
\textit{École Nationale Supérieure d'Informatique (ESI)}\\
Algiers, Algeria\\
Emails: mm\_mostefai@esi.dz, my\_tati@esi.dz, mi\_badaoui@esi.dz, mn\_bousdjira@esi.dz, ms\_hasnaoui@esi.dz, mr\_zitouni@esi.dz
}
}

\maketitle

\begin{abstract}
The one-dimensional \textit{Bin Packing Problem} (1D-BPP) consists of assigning a set of items with known sizes to a minimum number of identical bins of fixed capacity, without exceeding the capacity of any bin. This problem is NP-hard and quickly becomes intractable for exact methods as instance size grows. In this work, our central focus is a hybridization between the \textit{Hybrid Grouping Genetic Algorithm} (HGGA) metaheuristic and reinforcement learning. The proposed approach, denoted RL-HGGA, retains the classical operators of HGGA but replaces their fixed probabilistic selection with an adaptive policy learned via Q-learning. The agent observes the current state of the search, selects a macro-action, and updates its policy based on a reward derived from solution improvement and computational cost. Experiments are conducted on standard 1D-BPP benchmarks (Falkenauer, Scholl, and Hard28) and compare FFD, HGGA, and RL-HGGA in terms of number of bins used, gap to the lower bound, and execution time. Results show that HGGA remains the best reference in average solution quality, whereas RL-HGGA provides an interesting quality--time trade-off by substantially reducing computation time while maintaining quality close to that of HGGA. The ablation study shows that the ML component primarily improves the time efficiency of the search while preserving solution quality close to that of HGGA.
\end{abstract}


\section{Introduction}

\subsection{Problem Definition}
The one-dimensional \textit{Bin Packing Problem} is a grouping and assignment problem. Given a set of items $I=\{1,\ldots,n\}$, where each item $i$ has a size $w_i$, and a set of identical bins of capacity $C$, the goal is to assign each item to a bin such that the capacity is never exceeded, while minimizing the total number of bins used. This problem arises naturally in logistics, industrial cutting, memory allocation, file storage, and resource loading. Its theoretical difficulty is well established: the 1D-BPP is NP-hard, which explains the widespread use of approximate methods as instance size increases \cite{garey1979,johnson1974}.

\subsection{State of the Art of Metaheuristics}
The literature on solving the 1D-BPP can be organized into three main families. The first encompasses constructive heuristics, such as \textit{First Fit}, \textit{Best Fit}, \textit{First Fit Decreasing} (FFD), and \textit{Best Fit Decreasing}. These methods build a solution by scanning items in a given order and placing them in an existing bin or opening a new one. They are valued for their simplicity and speed, but remain limited by their greedy nature: a poor decision made early in construction is generally not revised afterward \cite{johnson1974}.

The second family comprises exact and near-exact methods, such as integer linear programming, branch-and-bound, and branch-and-price schemes. These approaches are important because they can prove optimality or provide reference bounds. However, their computational cost grows rapidly with instance size and difficulty. In an experimental context where multiple benchmark families must be processed, approximate methods therefore remain necessary.

The third family, central to this work, is that of metaheuristics. Metaheuristics are general-purpose search methods capable of exploring a very large solution space without enumerating it completely. They may be population-based, such as genetic algorithms, ant colony optimization, or swarm algorithms, or single-solution-based, such as tabu search, simulated annealing, or advanced local search methods. Their strength lies in combining \textbf{exploration}, to visit different regions of the search space, and \textbf{exploitation}, to locally improve promising solutions.

In a comparative study dedicated to the 1D-BPP, Munien \textit{et al.} compare several nature-inspired metaheuristic approaches, including variants of the Firefly Algorithm, Artificial Bee Colony, Cuckoo Search, a hybrid of Cuckoo Search with a genetic algorithm, and a classical genetic algorithm \cite{munien2020}. Their study shows that metaheuristics can obtain good-quality solutions on instances of varying difficulty, but also that computation time and solution quality depend heavily on the chosen algorithm, the internal heuristic used, and the structure of the instances. This observation confirms that solving the 1D-BPP is not merely a matter of choosing a high-performing metaheuristic on average: its behavior must also be controlled according to the current state of the search.

For the specific case of the BPP, an important advance was the \textit{Grouping Genetic Algorithm} (GGA), followed by its hybrid version HGGA proposed by Falkenauer \cite{falkenauer1996}. Unlike a classical genetic algorithm, which would represent a solution as a permutation of items, the GGA directly represents a solution as a set of groups, each group corresponding to a bin. This representation is more natural for the BPP, since the essential information is not only the order of items but the way in which they are grouped into bins.

Classical metaheuristics nonetheless exhibit several limitations. First, they depend heavily on parameter tuning: population size, mutation rate, local search probability, diversification intensity, stopping criterion, and so on. Second, operator application probabilities are often fixed, even though a given operator may not be equally useful at all stages of the search. For instance, a strong mutation may be beneficial when the search stagnates, but wasteful or costly when the population is already improving. Finally, metaheuristics may suffer from premature convergence, diversity loss, or high computational cost. These limitations motivate the addition of an adaptive mechanism capable of steering the search according to its current state.

\subsection{Scientific Motivation for Hybridization with ML}
The idea of hybridizing a metaheuristic with \textit{Machine Learning} (ML) rests on a simple observation: many internal decisions in an optimization algorithm are still governed by fixed rules or static probabilities. Yet during execution, the state of the search evolves: current solution quality, population diversity, stagnation, proximity to a lower bound, or operator cost. It is therefore natural to seek a mechanism capable of dynamically adapting algorithmic decisions to this current state.

Several hybridization modes are possible: learning parameter settings, operator selection, neighborhood choice, learning a diversification policy, or predicting a resolution profile from instance features. In our case, we retain operator selection via reinforcement learning, since this decision recurs at every generation and directly influences the exploration--exploitation trade-off. Q-learning is particularly well suited to this type of sequential decision problem: the agent observes a state, selects an action, receives a reward, and progressively adjusts its policy \cite{sutton2018,watkins1992}.

\subsection{State of the Art of ML--Metaheuristic Hybridization}
The hybridization of ML and metaheuristics falls within a broader framework, that of \textit{hyper-heuristics} and learning-assisted optimization methods. A first line of work concerns \textit{selection hyper-heuristics}, where a high-level mechanism dynamically selects the heuristic or operator to apply from a given portfolio. This idea has demonstrated its value across many combinatorial problems, as it replaces rigid strategies with adaptive policies \cite{burke2013}.

A second line of work concerns the use of reinforcement learning to drive local decisions within an optimization algorithm. The agent learns, through repeated interactions, which actions are most rewarding depending on the phase of the search. This principle has been explored in several NP-hard problems, including scheduling, routing, and combinatorial optimization more broadly, where it is used to select neighborhoods, priority rules, or operators \cite{bengio2021,khalil2017}.

In the specific case of the 1D-BPP, the literature remains richer in classical heuristics and metaheuristics than in explicitly learning-guided approaches. This reinforces the interest of a study such as ours: rather than modifying the structure of HGGA, we seek to preserve its effective operators and make only their selection mechanism adaptive. This strategy allows us to better isolate the contribution of the ML component while remaining faithful to a well-established reference metaheuristic for the BPP \cite{falkenauer1996}.

\subsection{Contribution and Originality}
The objective of this work is not to propose an entirely new metaheuristic, but to design a targeted, coherent, and analyzable hybridization between HGGA and reinforcement learning. The main contributions are as follows:
\begin{itemize}
    \item proposing an \textbf{RL-HGGA} variant in which the ML component is inserted at the critical decision point of the algorithm, namely the selection of the operator to apply at each generation;
    \item preserving the backbone of HGGA (group encoding, BPCX crossover, mutation, local search, selection, and replacement), so that the effect of the intelligent component can be isolated and interpreted;
    \item defining a \textbf{search state} composed of informative descriptors of the population and the progress of the algorithm;
    \item defining an \textbf{action space} in the form of macro-actions associating HGGA operators with varying search intensities;
    \item justifying the \textbf{placement of the ML component}: the agent is positioned inside the evolutionary loop, where the decision must be made repeatedly, contextually, and in a manner dependent on the current behavior of the population;
    \item conducting an \textbf{ablation study} by comparing HGGA without ML against RL-HGGA, in order to assess the actual contribution of the Q-learning component.
\end{itemize}

The originality of this work therefore lies less in the invention of new operators than in the way they are orchestrated: instead of fixed control, RL-HGGA learns a context-aware application policy, which makes it possible to explicitly analyze the contribution of a Q-learning agent over a grouping metaheuristic dedicated to the BPP.

\subsection{Paper Organization}
Section~II formalizes the 1D-BPP. Section~III presents the proposed RL-HGGA approach, its overall architecture, its encoding, and the Q-learning component. Section~IV describes the experimental protocol, datasets, hardware and software environment, and the parameter settings adopted. Section~V analyzes the experimental results and the actual effect of the hybridization. Finally, Section~V concludes the paper and outlines future directions.

\section{Mathematical Formulation of the Problem}

\subsection{Formal Definition}
Let $I=\{1,2,\ldots,n\}$ be a set of items, where each item $i$ has a positive size $w_i$. Items must be packed into identical bins of capacity $C$. The goal is to minimize the number of bins used. Since any item can always be placed alone in a bin, a trivial upper bound on the number of bins is $U=n$.

\subsection{Decision Variables}
Two types of binary variables are defined:
\begin{equation}
x_{ij}=\begin{cases}
1, & \text{if item } i \text{ is placed in bin } j,\\
0, & \text{otherwise,}
\end{cases}
\end{equation}
with $i\in\{1,\ldots,n\}$ and $j\in\{1,\ldots,U\}$.

We also define:
\begin{equation}
y_j=\begin{cases}
1, & \text{if bin } j \text{ is used,}\\
0, & \text{otherwise.}
\end{cases}
\end{equation}

\subsection{Objective Function}
The objective function minimizes the total number of bins used:
\begin{equation}
\min Z = \sum_{j=1}^{U} y_j.
\end{equation}

\subsection{Constraints}
Each item must be assigned to exactly one bin:
\begin{equation}
\sum_{j=1}^{U} x_{ij}=1, \quad \forall i\in\{1,\ldots,n\}.
\end{equation}

The capacity of each bin must not be exceeded:
\begin{equation}
\sum_{i=1}^{n} w_i x_{ij} \leq C y_j, \quad \forall j\in\{1,\ldots,U\}.
\end{equation}

Variables are binary:
\begin{equation}
x_{ij}\in\{0,1\}, \quad y_j\in\{0,1\}.
\end{equation}

An optional symmetry-breaking constraint may be added to avoid equivalent solutions obtained by permuting bins:
\begin{equation}
y_j \geq y_{j+1}, \quad \forall j\in\{1,\ldots,U-1\}.
\end{equation}

\subsection{Lower Bound Used in Evaluation}
A natural lower bound is given by:
\begin{equation}
LB=\left\lceil \frac{\sum_{i=1}^{n} w_i}{C}\right\rceil.
\end{equation}
In the experiments, this bound is used to normalize evaluation and compute a relative gap between methods:
\begin{equation}
Gap(\%) = \frac{Bins_{method}-LB}{LB}\times 100.
\end{equation}

\section{Description of the Proposed Hybrid Approach}

\subsection{Overall Architecture}
The proposed solution builds on a base metaheuristic, HGGA, subsequently steered by reinforcement learning. A BPP instance is first transformed into an initial population of feasible solutions using a noise-augmented variant of FFD. This population is then improved by an HGGA-style evolutionary loop: parent selection, application of variation and improvement operators, solution evaluation, and replacement of the weakest individuals.

The essential difference between classical HGGA and RL-HGGA lies in the choice of operators. In HGGA, the choice of crossover, mutation, or local search is governed by probabilities fixed in advance. In RL-HGGA, this choice is delegated to a Q-learning agent. At each generation, the agent observes the state of the search, selects a macro-action, and receives a reward based on the effect on population quality and computational cost.

\begin{figure*}[!t]
\centering
\resizebox{\textwidth}{!}{%
\begin{tikzpicture}[
    >=Stealth,
    bbox/.style={draw=white, rounded corners=5pt, align=center, text=white,
                 font=\small\bfseries, inner sep=5pt, minimum height=1.55cm},
    gbox/.style={bbox, fill=gray!58,          minimum width=1.9cm},
    tbox/.style={bbox, fill=teal!72!black,    minimum width=2.3cm},
    pbox/.style={bbox, fill=violet!55!blue,   minimum width=2.3cm},
    mbox/.style={bbox, fill=magenta!70!black, minimum width=2.3cm, minimum height=2.1cm},
    qbox/.style={bbox, fill=purple!62,        minimum width=2.3cm, minimum height=2.1cm},
    obox/.style={bbox, fill=orange!75!red,    minimum width=2.3cm, minimum height=2.1cm},
    sa/.style={->, thick},
    da/.style={->, thick, dashed, black},
    ra/.style={->, thick, dashed, red!80},
]

\node[gbox] (INST) at ( 0.0, 0)
    {BPP Instance\\\footnotesize$(n,\,C,\,w_1\!\ldots w_n)$};
\node[tbox]  (INIT) at ( 3.4, 0)
    {Initialization\\\footnotesize population\\random-FFD};
\node[tbox]  (POP)  at ( 6.8, 0)
    {Current Population\\\footnotesize chromosomes\\$=$ grouped bins};
\node[pbox]  (MAC)  at (10.4, 0)
    {Macro-action\\\footnotesize$A_0\!\ldots A_7$\\BPCX, mutation,\\local search, restart};
\node[tbox]  (EVL)  at (13.9, 0)
    {Evaluation\\\footnotesize Falkenauer fitness\\update\\best individual};
\node[gbox, minimum width=0.75cm, minimum height=2.6cm] (OUT) at (16.8, 0)
    {\rotatebox{90}{\textbf{Output}}};

\draw[sa] (INST)--(INIT); \draw[sa] (INIT)--(POP);
\draw[sa] (POP)--(MAC);   \draw[sa] (MAC)--(EVL);
\draw[sa] (EVL)--(OUT);

\node[mbox] (STA) at ( 6.8, -5.0)
    {\textbf{State} $s_t$\\\footnotesize progress\\stagnation, gap,\\avg.\ fitness\\variance,\\fill rate};
\node[qbox] (AGT) at (10.4, -5.0)
    {\textbf{Q-learning Agent}\\\footnotesize policy\\$\varepsilon$-greedy\\Q-table $Q(s,a)$};
\node[obox] (REW) at (13.9, -5.0)
    {\textbf{Reward}\\\footnotesize solution quality\\improvement\\$+$ CPU cost};

\draw[sa] (STA)--(AGT); \draw[sa] (AGT)--(REW);


\draw[da] (POP.south)
    -- node[left, font=\footnotesize, xshift=-3pt] {state}
    (STA.north);

\draw[da] (AGT.north)
    -- node[right, font=\footnotesize, xshift=3pt] {action $a_t$}
    (MAC.south);

\draw[da] (EVL.south) -- (REW.north);

\coordinate (NE) at ($(EVL.north)+(0,0.9)$);
\coordinate (NP) at ($(POP.north)+(0,0.9)$);
\draw[ra] (EVL.north) -- (NE)
    -- node[above, font=\footnotesize, text=red!80] {new generation}
    (NP) -- (POP.north);

\draw[sa] (REW.south) -- ++(0,-0.4)
    node[below right, font=\footnotesize, xshift=3pt] {update $Q(s_t,a_t)$}
    -| (AGT.south);

\begin{pgfonlayer}{background}
    \node[draw=gray!25, fill=white, rounded corners=8pt, inner sep=9pt,
          fit=(INST)(OUT)(EVL)(POP)(MAC), name=TF] {};
    \node[draw=gray!25, fill=white, rounded corners=8pt, inner sep=9pt,
          fit=(STA)(AGT)(REW), name=BF] {};
\end{pgfonlayer}

\node[font=\small\bfseries, text=gray!55, rotate=90]
    at ($(TF.west)+(-0.45,0)$) {Main Flow};
\node[font=\small\bfseries, text=gray!55, rotate=90]
    at ($(BF.west)+(-0.45,0)$) {ML Controller};

\end{tikzpicture}%
}
\caption{Detailed architecture of RL-HGGA and placement of the ML component within the evolutionary loop.}
\label{fig:architecture_rl_hgga}
\end{figure*}
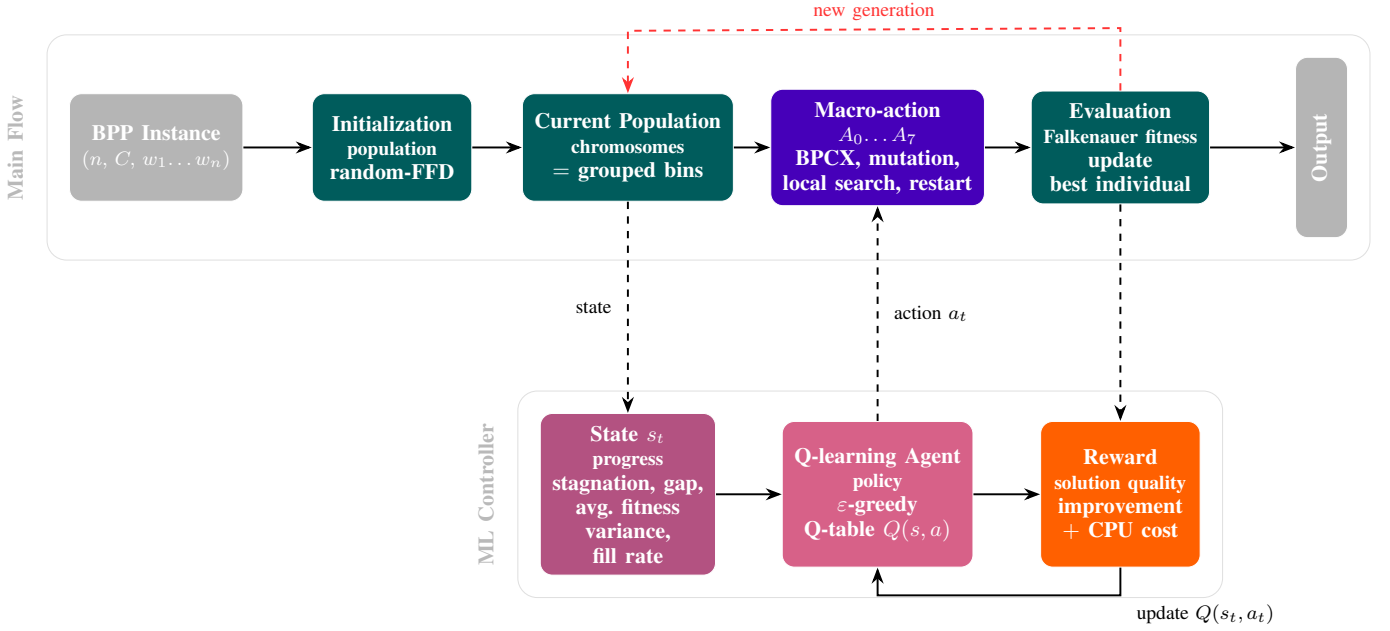

Fig.~\ref{fig:architecture_rl_hgga} shows the final architecture diagram. It clearly illustrates the main HGGA loop and the insertion point of the ML component: the Q-learning agent intervenes solely to drive operator selection, without replacing the HGGA operators themselves.

\subsection{Detailed Principles of HGGA}
HGGA stands for \textit{Hybrid Grouping Genetic Algorithm}. It is an adaptation of genetic algorithms to grouping problems. In a classical genetic algorithm, a population of solutions evolves through operators such as selection, crossover, and mutation. This general scheme is retained in HGGA but is modified to better match the structure of the Bin Packing Problem.

The distinctive feature of HGGA is that the key unit is not the individual item but the \textbf{group of items}. In the BPP, a group corresponds to a bin. Thus, the algorithm's objective is to evolve item groupings and to preserve well-constructed bins across generations. This idea is essential: if a bin is nearly full without exceeding capacity, it represents a good structure that should be preserved during crossover.

The general operation of HGGA can be summarized in four steps. First, an initial population of feasible solutions is generated. Second, parents are selected, typically by tournament, to favor high-quality individuals without entirely eliminating diversity. Third, specialized operators produce new individuals: BPCX crossover transfers entire groups from one parent to another, mutation dissolves certain bins and reinserts their items, and local search improves groupings through item moves or swaps. Fourth, a $(\mu+\lambda)$ replacement retains the best individuals among parents and offspring.

HGGA is thus a hybrid metaheuristic for two reasons. On one hand, it combines a global evolutionary logic, exploring multiple regions of the search space through a population. On the other hand, it integrates local improvement and repair mechanisms, which exploit the best solutions already found. This combination yields better solutions than a simple constructive heuristic such as FFD, but also increases computation time. It is precisely this limitation that motivates adding a Q-learning component in RL-HGGA.

\subsection{Encoding and Solution Representation}
The encoding implemented in the notebooks closely follows the logic of the \textit{Grouping Genetic Algorithm}. A solution is not represented as a simple permutation of items, but as a \textbf{chromosome} consisting of an ordered list of \textbf{groups}. Each group corresponds to a bin.

In the implementation, a \texttt{Group} contains a list of \texttt{(item\_id, weight)} pairs and a \texttt{fill} attribute representing the total load of the bin. This structure allows immediate feasibility testing of an insertion via the condition $fill + w \leq C$. A \texttt{Chromosome} is then defined as a list of groups, i.e., a complete packing.

This representation offers several practical and algorithmic advantages.
\begin{itemize}
    \item It is \textbf{natural} for the BPP, since the problem consists precisely of partitioning items into bins.
    \item It facilitates the \textbf{preservation of good groups} during crossover, unlike a permutation encoding where grouping information is implicit.
    \item It makes \textbf{repair and local improvement} simpler, since items can be directly moved, swapped, or reinserted between groups.
\end{itemize}

The quality of a chromosome is evaluated using Falkenauer's fitness function:
\begin{equation}
fitness(X)=\frac{1}{|X|}\sum_{g\in X}\left(\frac{fill(g)}{C}\right)^k,
\end{equation}
with $k=2$ in the implementation. This function rewards well-filled bins and penalizes dispersed fillings, which is consistent with the objective of minimizing the number of bins \cite{falkenauer1996}.

\subsection{Population Initialization}
The initial population is not generated uniformly at random. The notebooks use a \textit{random-FFD} variant: items are first shuffled, then partially sorted in blocks before applying \textit{First Fit}. This introduces diversity while maintaining good initial quality. In practice, each initial individual is a feasible solution that is relatively competitive but distinct from the others. This step is important, as it provides HGGA with a starting population that is both varied and already structured.

\subsection{HGGA Operators Preserved in RL-HGGA}
RL-HGGA preserves all HGGA operators exactly; only their selection mechanism changes.

\textbf{BPCX Crossover.} The \textit{Bin Packing Crossover} (BPCX) selects a segment of groups from one parent, injects groups from the second parent, removes item duplications, and reinserts orphan items using an FFD-like mechanism. The goal is to preserve high-quality bins rather than systematically breaking up groupings.

\textbf{Light and Heavy Mutation.} Mutations are implemented through partial dissolution of bins followed by greedy reinsertion. Light mutation dissolves a small fraction of underfilled bins, while heavy mutation dissolves a larger fraction. In the code, this logic is realized through \texttt{dissolve\_and\_repack}, with a low or high fraction parameter respectively.

\textbf{Local Search.} The local search used follows a Martello--Toth style. The least-filled bins are examined first, and the algorithm attempts dominance moves such as replacing two items with one or vice versa, in order to eliminate inefficient bins and densify the grouping \cite{martello1990}.

\textbf{Selection and Replacement.} Parents are chosen by tournament, and replacement follows a $(\mu+\lambda)$ strategy: parents and offspring are pooled, sorted by fitness, and only the best individuals are retained. This strategy maintains stable selection pressure while avoiding the immediate loss of good solutions.

\subsection{Q-learning Component: States, Actions, and Reward}
The Q-learning agent is the core of the intelligent module. At each generation, it receives a discretized state of the search. This state consists of eight normalized continuous features, subsequently discretized to feed a tabular Q-table:
\begin{itemize}
    \item generation progress;
    \item stagnation level;
    \item gap of the best solution with respect to the lower bound;
    \item average population fitness;
    \item fitness variance;
    \item fraction of nearly-full bins;
    \item fraction of underfilled bins;
    \item recent improvement slope.
\end{itemize}

The action space consists of eight \textbf{macro-actions}:
\begin{itemize}
    \item A0: BPCX + light mutation;
    \item A1: BPCX + light mutation + local search;
    \item A2: BPCX + heavy mutation;
    \item A3: BPCX + heavy mutation + local search;
    \item A4: BPCX + local search;
    \item A5: local search only on the best individual;
    \item A6: restart of the worst individuals;
    \item A7: mutation only.
\end{itemize}

The choice of these actions is important: the agent does not learn an abstract elementary operator, but directly selects a \textbf{search recipe} corresponding to a given style of evolution. This makes learning more interpretable.

For the main RL-HGGA variant, the reward combines the improvement in best fitness, the improvement in average fitness, and a penalty related to the CPU cost of the action. The update rule is that of classical Q-learning:
\begin{equation}
Q(s,a) \leftarrow Q(s,a)+\alpha\left[r+\gamma\max_{a'}Q(s',a')-Q(s,a)\right].
\end{equation}
The design of this reward is central, as it determines the trade-off between solution improvement and the temporal cost of operators.

\begin{algorithm}[htbp]
\caption{General scheme of RL-HGGA}
\begin{algorithmic}[1]
\STATE Generate initial population using \textit{random-FFD}
\STATE Initialize Q-learning agent
\FOR{each generation}
    \STATE Extract current state $s_t$
    \STATE Select action $a_t$ using $\epsilon$-greedy policy
    \STATE Apply HGGA macro-action corresponding to $a_t$
    \STATE Evaluate new population and update best individual
    \STATE Compute reward $r_t$
    \STATE Observe new state $s_{t+1}$
    \STATE Update $Q(s_t,a_t)$
\ENDFOR
\STATE Return best chromosome obtained
\end{algorithmic}
\end{algorithm}

\section{Experiments, Results, and Analysis}

\subsection{Experimental Protocol}

The experimental protocol aims to evaluate the contribution of reinforcement learning within the HGGA metaheuristic for the one-dimensional Bin Packing Problem. Three main methods are compared: FFD, used as a fast heuristic baseline; HGGA, used as a metaheuristic baseline without an intelligent component; and RL-HGGA, which constitutes the proposed hybrid approach.

The test scenario is organized in two phases. In the first phase, the Q-learning agent of RL-HGGA is trained on a diverse set of instances in order to learn an operator selection policy. During this phase, the agent selects actions using an $\epsilon$-greedy strategy, balancing exploration and exploitation. In the second phase, the methods are evaluated on the selected benchmark families. For RL-HGGA, evaluation uses the learned greedy policy, meaning the agent selects at each generation the action with the highest $Q(s,a)$ value for the current state.

All methods are run on the same instance families and compared using the same metrics: average number of bins used, average gap to the lower bound, and average execution time per instance. This setup allows FFD, HGGA, and RL-HGGA to be compared under homogeneous conditions and enables analysis of the quality--time trade-off. An ablation study is also conducted by comparing HGGA without ML against RL-HGGA, in order to assess the actual contribution of the Q-learning component.

\begin{table}[htbp]
\centering
\caption{Hardware and software environment for the experiments.}
\label{tab:env}
\begin{tabular}{p{2.5cm}p{5.2cm}}
\toprule
Component & Configuration \\
\midrule
Machine & ASUS Vivobook X7600PC\_N7 \\
Operating System & Ubuntu 24.04.4 LTS x86\_64, kernel 6.17.0-35-generic \\
CPU & Intel Core i7-11370H, 8 logical threads \\
Memory & 16 GiB RAM \\
Runtime Environment & Jupyter Notebook / Python \\
Main Libraries & NumPy, Pandas, Matplotlib; scikit-learn used when necessary \\
\bottomrule
\end{tabular}
\end{table}

\subsection{Datasets}
The experiments use classical 1D-BPP benchmarks, in particular the Falkenauer (T and U), Scholl, and Hard28 families. These families cover varied instance sizes and different weight structures, enabling evaluation of average quality, robustness, and the computational scaling behavior of the methods as the number of items increases \cite{delorme2018}.

\begin{table}[htbp]
\centering
\caption{Benchmark families used.}
\label{tab:benchmarks}
\begin{tabular}{lccc}
\toprule
Family & No. inst. & Size $n$ & Purpose \\
\midrule
Falkenauer T60 & 20 & 60 & Quality / small scale \\
Falkenauer T120 & 20 & 120 & Quality / medium scale \\
Falkenauer T249 & 20 & 249 & Quality / medium scale \\
Falkenauer T501 & 20 & 501 & Large scale \\
Falkenauer U120--U1000 & 80 & 120--1000 & Robustness \\
Scholl 1--3 & variable & 50--500 & Instance diversity \\
Hard28 & 28 & 160--200 & Hard instances \\
\bottomrule
\end{tabular}
\end{table}

\subsection{Parameter Calibration}

HGGA and RL-HGGA are evaluated with the same main parameters to ensure a fair comparison. The values retained are $\mu=25$, \textit{max\_gen}$=80$, and \textit{patience}$=25$. These parameters directly control the trade-off between solution quality and computation time.

The parameter $\mu$ denotes the population size, i.e., the number of chromosomes maintained at each generation. With $\mu=25$, the algorithm keeps 25 candidate solutions in parallel. A larger population can increase diversity and reduce the risk of premature convergence, but also increases execution time, as more solutions must be crossed, mutated, and evaluated at each generation.

The parameter \textit{max\_gen} represents the maximum number of generations allowed. One generation corresponds to a complete iteration of the evolutionary loop: parent selection, operator application, offspring generation, evaluation, and replacement. With \textit{max\_gen}$=80$, the algorithm may perform at most 80 improvement cycles for a given instance. This parameter acts as a search budget: a higher value gives the algorithm more time to improve solutions but increases computational cost.

The parameter \textit{patience} is an early stopping criterion based on stagnation. With \textit{patience}$=25$, if the best solution does not improve for 25 consecutive generations, execution stops before reaching \textit{max\_gen}. This mechanism avoids continuing the search unnecessarily when the population appears stuck at the same solution quality. It thus reduces execution time without imposing an overly premature halt.

For training RL-HGGA, the Q-learning agent is trained over 400 episodes. Each episode corresponds to one execution of the RL-HGGA loop on a training instance. The agent uses an $\epsilon$-greedy strategy: initially, $\epsilon$ is high to encourage exploration of the different macro-actions; it then decays progressively to 0.05 to favor actions that have achieved the best values in the $Q(s,a)$ table. This schedule allows the agent to explore multiple behaviors at the start of training, then progressively exploit the learned policy.

\subsection{Main Results: Solution Quality}

\begin{figure*}[!t]
\centering
\includegraphics[width=0.96\textwidth]{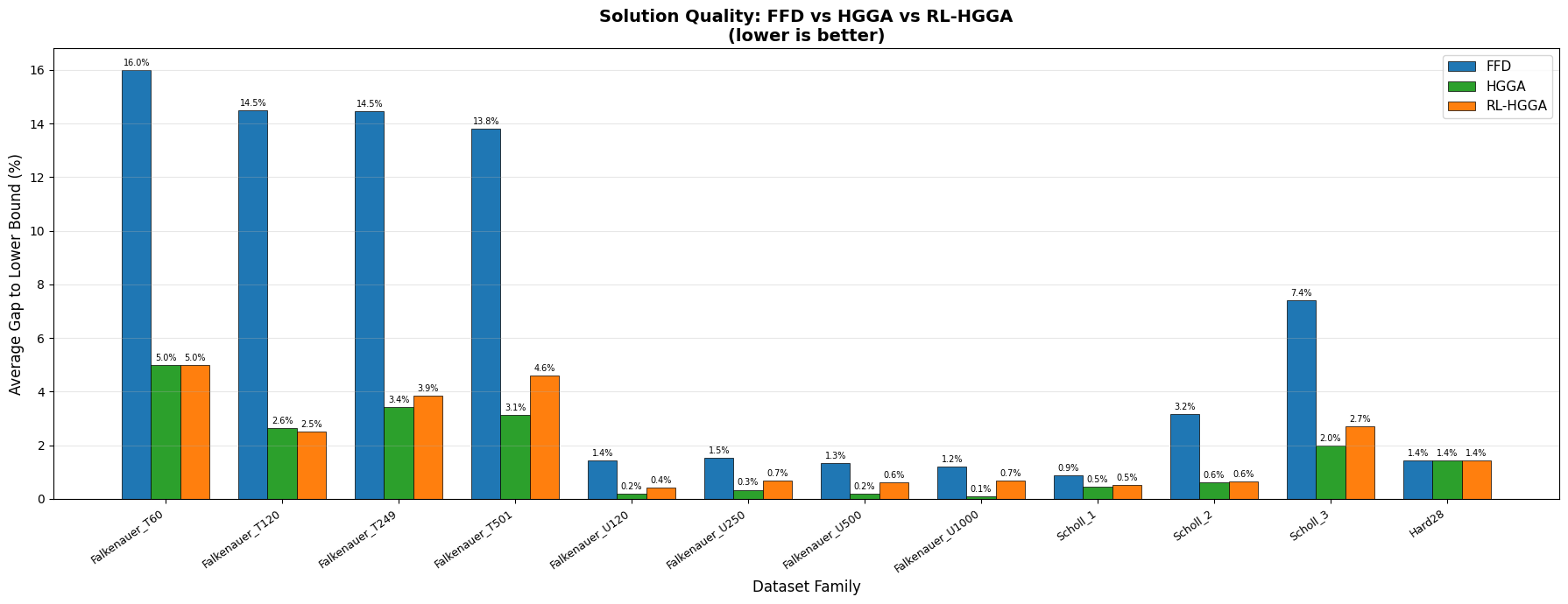}
\caption{Average gap to the lower bound per family: comparison between FFD, HGGA, and RL-HGGA. Lower values indicate better solutions.}
\label{fig:quality_gap_main}
\end{figure*}

Solution quality is first analyzed through the average gap to the lower bound. Fig.~\ref{fig:quality_gap_main} compares FFD, HGGA, and RL-HGGA across the different instance families. FFD exhibits the highest gaps, particularly on the Falkenauer T families. This result confirms the limitation of greedy heuristics: they produce a feasible solution very quickly but lack any improvement mechanism capable of reorganizing bins after the initial decisions.

HGGA generally achieves the best average gaps. This is explained by its ability to explore multiple groupings through BPCX crossover, mutations, and local search. RL-HGGA remains clearly better than FFD and often approaches HGGA, but does not systematically outperform HGGA in pure quality. This point is important for interpreting the results: the contribution of the ML component is not primarily reflected in an absolute improvement in the number of bins, but in better management of the quality--time trade-off.

\begin{figure*}[!t]
\centering
\includegraphics[width=0.96\textwidth]{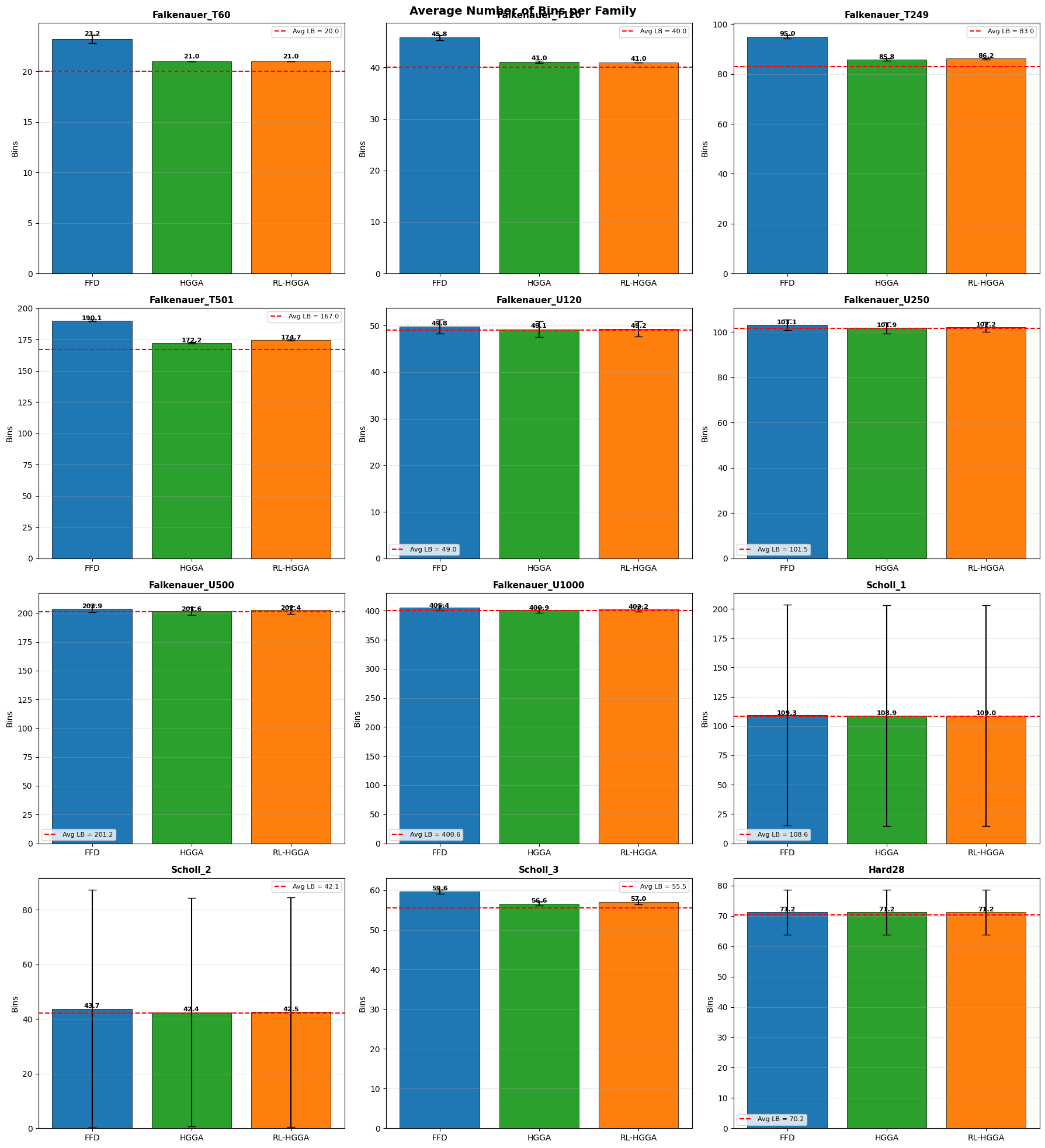}
\caption{Average number of bins used per family: comparison between FFD, HGGA, and RL-HGGA.}
\label{fig:avg_bins_main}
\end{figure*}

Fig.~\ref{fig:avg_bins_main} complements this analysis by comparing the average number of bins used by each method. Results confirm the trend observed with the gap: FFD generally uses more bins, while HGGA and RL-HGGA produce more compact packings. Across several families, RL-HGGA achieves a number of bins close to that of HGGA, showing that the Q-learning agent maintains good solution quality despite a faster search strategy.

\subsection{Main Results: Execution Time}

\begin{figure*}[!t]
\centering
\includegraphics[width=0.96\textwidth]{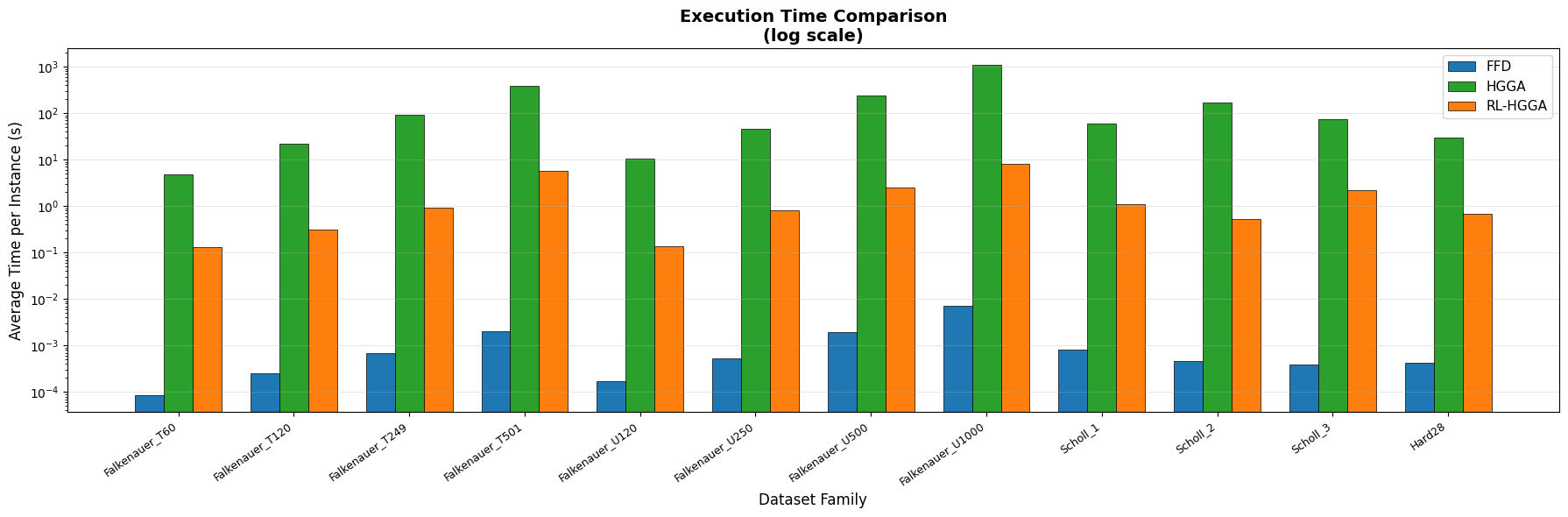}
\caption{Comparison of average execution times per family between FFD, HGGA, and RL-HGGA. The logarithmic scale allows visualization of large time differences.}
\label{fig:time_main}
\end{figure*}

Fig.~\ref{fig:time_main} presents average execution times per family on a logarithmic scale. FFD is logically the fastest method, as it applies a greedy construction without any improvement phase. HGGA is much more costly, as it maintains a population of solutions and applies evolutionary operators over multiple generations.

RL-HGGA falls between these two extremes, but is much closer to FFD than to HGGA in execution time. This shows that the Q-learning agent learns to favor less costly operator sequences while maintaining acceptable solution quality. The main result of RL-HGGA is therefore a quality--time trade-off: the method does not always improve upon HGGA in pure quality, but it substantially reduces computation time relative to HGGA.

\begin{table}[htbp]
\centering
\caption{Overall solution quality: aggregated values across all benchmark families.}
\label{tab:global_quality}
\begin{tabular}{lccc}
\toprule
Method & Avg. Gap & Med. Gap & Max Gap \\
\midrule
FFD & 2.47\% & 0.65\% & 20.00\% \\
HGGA & 0.75\% & 0.00\% & 14.29\% \\
RL-HGGA & 0.95\% & 0.00\% & 14.29\% \\
\bottomrule
\end{tabular}
\end{table}

\begin{table}[htbp]
\centering
\caption{Overall execution time: average time per instance aggregated across all benchmark families.}
\label{tab:global_time}
\begin{tabular}{lc}
\toprule
Method & Avg. time/inst. \\
\midrule
FFD & 0.0008 s \\
HGGA & 64.22 s \\
RL-HGGA & 1.29 s \\
\bottomrule
\end{tabular}
\end{table}

Tables~\ref{tab:global_quality} and~\ref{tab:global_time} summarize the quality--time trade-off by aggregating results across all benchmark families. Table~\ref{tab:global_quality} presents the global quality metrics: average gap, median gap, and maximum gap. HGGA achieves the best overall average gap at 0.75\%, compared to 2.47\% for FFD and 0.95\% for RL-HGGA. Both HGGA and RL-HGGA achieve a median gap of zero, indicating that on a significant portion of instances both methods reach the lower bound used. Their maximum gap is also identical at 14.29\%, while FFD reaches a higher maximum gap of 20.00\%.

Table~\ref{tab:global_time} shows, however, that RL-HGGA substantially reduces the average execution time relative to HGGA. The average time per instance drops from 64.22 seconds for HGGA to 1.29 seconds for RL-HGGA, while FFD naturally remains the fastest method at 0.0008 seconds per instance. Thus, the contribution of Q-learning is not to replace HGGA at peak quality, but to make the search far more time-efficient while preserving close quality.

\subsection{Analysis of RL-HGGA Behavior}

To understand the behavior of RL-HGGA, we analyze its training phase, the learned policy, and its convergence behavior.

\begin{figure}[htbp]
\centering
\includegraphics[width=\columnwidth]{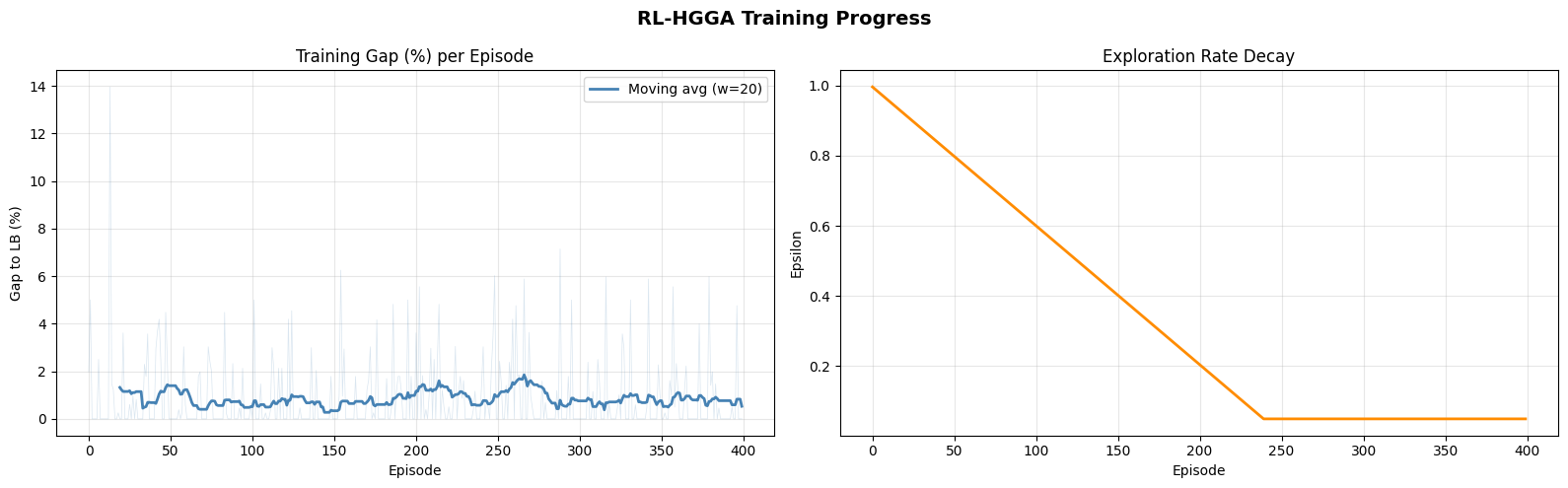}
\caption{RL-HGGA training progression: evolution of the training gap and decay of the exploration rate $\epsilon$.}
\label{fig:training}
\end{figure}

Fig.~\ref{fig:training} shows the evolution of the gap during training and the decay of the exploration rate $\epsilon$. At the beginning of training, the agent strongly explores the available actions. Progressively, $\epsilon$ decreases, meaning the agent increasingly exploits the actions with the best values in the $Q(s,a)$ table. The training gap remains variable, which is expected since episodes cover different instances, but the average curve indicates that the agent learns an exploitable policy.

\begin{figure}[htbp]
\centering
\includegraphics[width=\columnwidth]{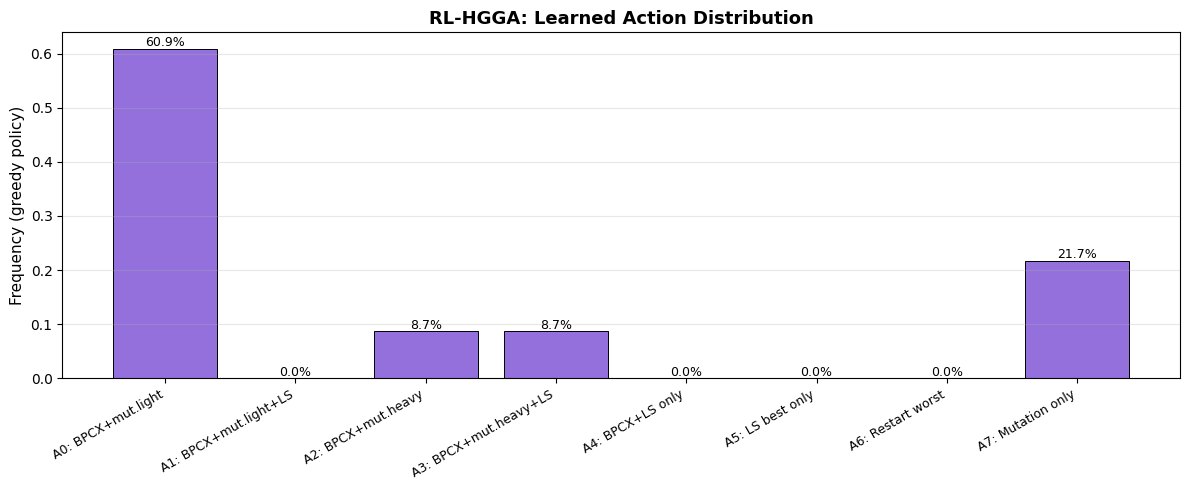}
\caption{Distribution of macro-actions learned by RL-HGGA under the greedy policy.}
\label{fig:actions}
\end{figure}

Fig.~\ref{fig:actions} shows the action distribution learned by the greedy policy. The agent does not select all actions uniformly. It primarily favors certain macro-actions, notably combinations based on BPCX with light mutation, as well as mutation alone. This indicates that the agent has learned to avoid certain more costly or less rewarding actions under the reward function used.

\begin{figure}[htbp]
\centering
\includegraphics[width=\columnwidth]{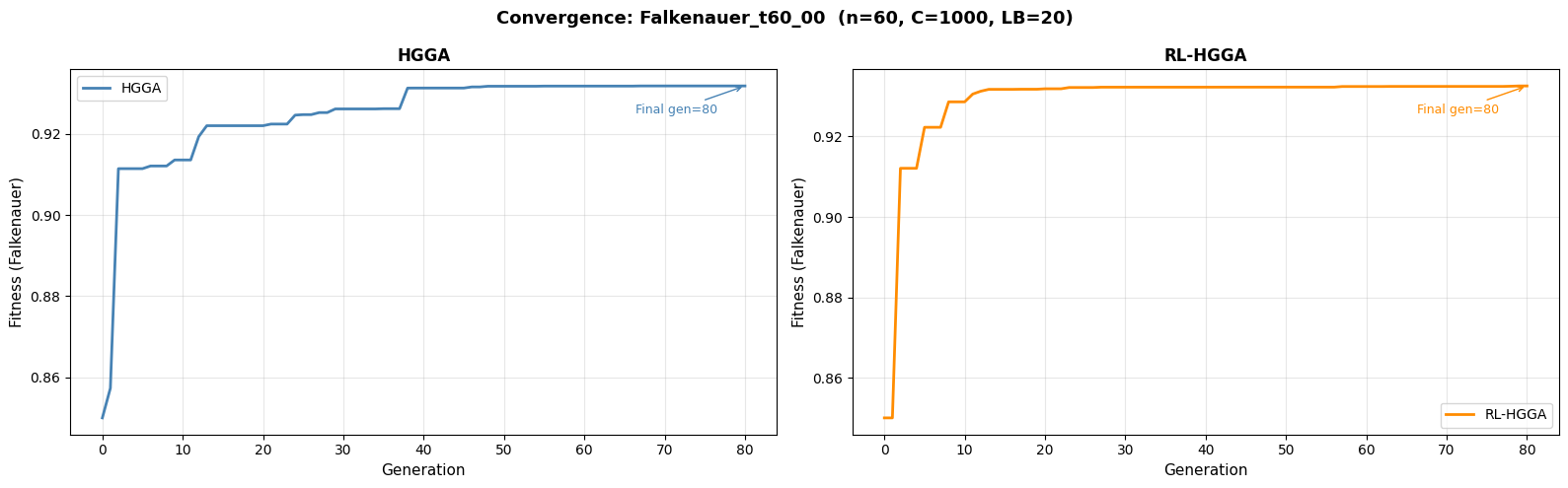}
\caption{Comparison of convergence behavior between HGGA and RL-HGGA on a reference instance.}
\label{fig:convergence_example}
\end{figure}

Fig.~\ref{fig:convergence_example} compares the convergence behavior of HGGA and RL-HGGA on a reference instance. RL-HGGA reaches good quality rapidly, confirming its orientation toward a more time-efficient search. HGGA may retain a slight final quality advantage on some instances, but at the cost of a much higher execution time.

\subsection{Ablation Study: Actual Contribution of the ML Component}

The ablation study consists of comparing HGGA without the ML component against RL-HGGA, which retains the same operators but replaces fixed operator selection with a Q-learning agent. This comparison isolates the effect of the learning module.

Results show that adding Q-learning primarily affects the temporal behavior of the algorithm. HGGA achieves a better overall average gap at 0.75\%, while RL-HGGA achieves 0.95\%. However, RL-HGGA substantially reduces the average execution time, dropping from 64.22 seconds for HGGA to 1.29 seconds per instance. The contribution of ML is therefore real, but must be interpreted correctly: it does not consist of systematically improving final solution quality, but of learning a faster search policy while maintaining quality close to that of HGGA.

From a critical standpoint, this ablation shows that the ML component brings a significant efficiency gain but also introduces a slight average quality loss. The design of the reward function is therefore critical. A reward too heavily oriented toward speed favors low-cost actions, while a reward too heavily oriented toward quality risks increasing computation time. The correct behavior of RL-HGGA thus depends on the balance between solution improvement and temporal cost.

\subsection{Comparative Study}
The comparative study quantitatively contrasts the methods implemented in this project: FFD, HGGA, and RL-HGGA. FFD represents the family of fast constructive heuristics. HGGA represents the grouping metaheuristic used as a reference without an intelligent component. RL-HGGA constitutes the proposed approach, in which reinforcement learning dynamically drives operator selection.

Results show that FFD is the best method in terms of speed but the weakest in solution quality. HGGA achieves the best overall average gap, but its execution time is much higher. RL-HGGA positions itself as an intermediate method: it maintains quality clearly better than FFD and close to HGGA, while substantially reducing execution time relative to HGGA.

Regarding comparison with the literature, it is important to note that no direct numerical confrontation with published values was conducted. Such a comparison would require the exact same instances, stopping criteria, hardware environment, and statistical protocol. In this work, the literature is therefore used as a \textbf{methodological reference} and scientific positioning. In particular, the study by Munien \textit{et al.} confirms that metaheuristics applied to the 1D-BPP exhibit a strong trade-off between solution quality and computation time \cite{munien2020}. Our contribution aligns with this: we compare our implemented methods against each other, then demonstrate that adding Q-learning is aimed precisely at better controlling this trade-off.

The comparative study must therefore be interpreted with care. It experimentally validates, within our protocol, that RL-HGGA improves time efficiency over HGGA while preserving close quality. It does not claim to demonstrate that RL-HGGA quantitatively outperforms all results in the literature, since that comparison was not conducted under strictly identical conditions.

\subsection{Discussion}
From the results, three conclusions emerge clearly.
\begin{itemize}
    \item \textbf{FFD} is indispensable as a speed baseline, but insufficient when solution quality is a priority.
    \item \textbf{HGGA} remains the reference in average quality; it is the best metaheuristic baseline among the methods studied.
    \item \textbf{RL-HGGA} is the main contribution of the project, as it provides a quality--time trade-off substantially more favorable than HGGA when computation cost must be controlled.
\end{itemize}

In other words, the ML component does not replace the heuristic expertise embodied in HGGA; it \textbf{steers} it. It is precisely this adaptive orchestration that constitutes the value of this work.

\section{Conclusion and Future Work}
This work proposed a hybridization of HGGA and reinforcement learning for the 1D-BPP. The RL-HGGA approach preserves the effective operators of the reference metaheuristic and makes their selection mechanism adaptive using a Q-learning agent. Experiments show that HGGA remains the best method in average solution quality, while RL-HGGA provides a highly favorable quality--time trade-off, with a significant reduction in execution time for a limited quality loss. The ablation study further confirms that the Q-learning component primarily brings a time efficiency gain, with a slight average quality loss relative to HGGA.

As future directions, several natural avenues arise: enriching the RL state with finer diversity indicators, refining the action space, learning not only the operator type but also its intensity, or replacing tabular Q-learning with deep reinforcement learning when the state space becomes too large. Another direction would be to integrate more general hyper-heuristic mechanisms to select not only operators but also certain algorithm parameters online.

\balance
\end{document}